\documentclass[11pt]{article}
\usepackage[margin=1in]{geometry}
\usepackage[T1]{fontenc}
\usepackage{lmodern,microtype,graphicx,booktabs,amsmath,enumitem,xcolor,array}
\usepackage[numbers,sort&compress]{natbib}
\usepackage[hypertexnames=false,colorlinks=true,linkcolor=teal,citecolor=teal,urlcolor=teal]{hyperref}
\usepackage{float,caption}
\setlist{nosep,leftmargin=*}
\title{Robot Manipulation with GPT-6-Astra:\\Body Knowledge, Experience Reuse,\\Emergent Skills, and Sim2Real Transfer}
\author{%
  \normalsize Sida He, Lingxi Xie, Yunning Cao, Pengfei Chen, Kaiwen Duan\\
  Jiannan Ge, Xinyue Huo, Jiacheng Shao, Qi Tian\textsuperscript{*}\\[3pt]
  \small Huawei Inc., China\\[2pt]
  \small\textsuperscript{*}Corresponding author
}
\date{September 2026}
\hypersetup{%
  pdftitle={Robot Manipulation with GPT-6-Astra: Body Knowledge, Experience Reuse, Emergent Skills, and Sim2Real Transfer},
  pdfauthor={Sida He, Lingxi Xie, Yunning Cao, Pengfei Chen, Kaiwen Duan, Jiannan Ge, Xinyue Huo, Jiacheng Shao, Qi Tian}
}
\makeatletter
\renewcommand{\maketitle}{%
  \begin{center}
  {\LARGE\@title\par}
  \ifx\@author\@empty\else\vspace{0.5em}{\large\@author\par}\fi
  \vspace{0.6em}{\large\@date\par}
  \vspace{0.3em}{\small\url{https://github.com/hesd10/astra-robot-sim2real}\par}
  \end{center}
  \vspace{0.2em}
}
\makeatother
\begin{document}
\begingroup
\setlength{\parskip}{3pt}
\maketitle
\begin{figure}[H]
\centering\includegraphics[width=.94\linewidth]{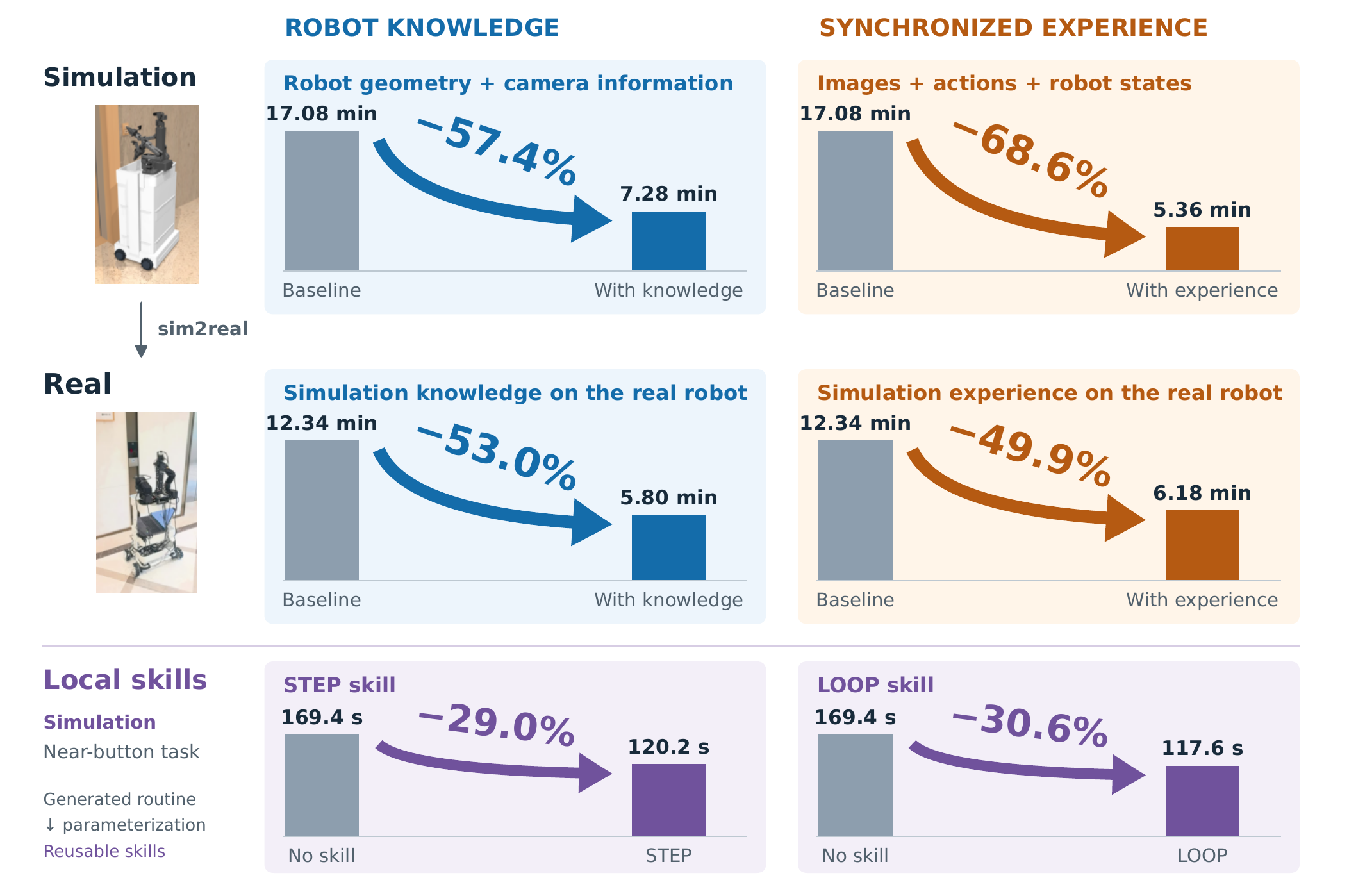}
\captionsetup{font=small}
\caption{\textbf{Faster elevator-button manipulation with GPT-6-Astra in simulation and on a real robot.} Robot geometry and camera information (blue), or synchronized images, actions, and robot states from a successful simulated attempt (orange), reduce task time relative to the same agent given neither resource. The sim2real arrow denotes reuse of simulation knowledge and experience on a calibrated physical robot. Bottom: researchers derived reusable skills from an agent-generated routine to accelerate the final approach in simulation. \textbf{STEP} performs one short movement and returns an image to the agent; \textbf{LOOP} repeats movements and image checks internally until the button's visual activation signal is detected or an execution limit is reached. Their baseline has no supplied skill. Bars show mean task time, including observation, model interaction, and execution (three trials per condition above; nine below). Percentages denote reductions relative to each baseline; bar pairs start at zero and are scaled independently. Images illustrate the task.}
\label{fig:teaser}
\end{figure}
\endgroup
\clearpage
\begin{abstract}
General-purpose multimodal agents can write robot-control programs, but repeated exploration and model-mediated action selection can make execution slow. We study how external body knowledge, successful experience, and executable skills improve an XLeRobot controlled by GPT-6-Astra in a simulated and a physical elevator-button task. In 30 fixed-start simulation trials, complete robot geometry and camera information reduce mean completion time by 57.4\% relative to a baseline with only the common control interface and no prior experience; images with synchronized action and state records reduce it by 68.6\% without additional body assets. In nine paired comparisons (18 trials) at starts displaced by 10--100\,cm, experience recorded at the original start reduces mean time by 58--63\% relative to no experience, demonstrating generalization to the tested new starting positions. During experience experiments, GPT-6-Astra spontaneously generates a short visual-feedback program. Researcher-refactored versions reduce mean local-task time by 29--31\% in 27 simulation trials. Finally, 12 real-robot trials using operator-confirmed button contact demonstrate sim2real reuse: at a shared nominal start, simulation XML assets and simulation experience reduce mean time by 53.0\% and 49.9\%, respectively; real experience also transfers to two new starts. These results suggest a practical way to build general-purpose manipulation experiments around GPT-6-Astra: supply machine-readable body descriptions and synchronized demonstrations, and turn useful agent-generated feedback routines into reusable skills, while the agent adapts actions from current images. We release all task prompts, trial-level experimental data, and acquired skill implementations at \url{https://github.com/hesd10/astra-robot-sim2real}.
\end{abstract}

\section{Introduction}
Sun Tzu, an ancient Chinese military strategist, wrote in \emph{The Art of War}: ``If you know the enemy and know yourself, you need not fear the result of a hundred battles.''~\cite{suntzu1910} For an embodied agent, knowing itself begins with understanding the robot it controls. GPT-6-Astra (Astra below) is a general-purpose multimodal coding agent connected to a robot through a software interface. We vary the body knowledge supplied to it, from joint names, limits, and readings to link dimensions, joint-axis geometry, and camera mounting. How much more efficiently can GPT-6-Astra act when it knows the robot's physical structure? We study this question alongside two related resources: records of successful attempts and reusable control routines.

\begin{figure}[t]
\centering\includegraphics[width=\linewidth]{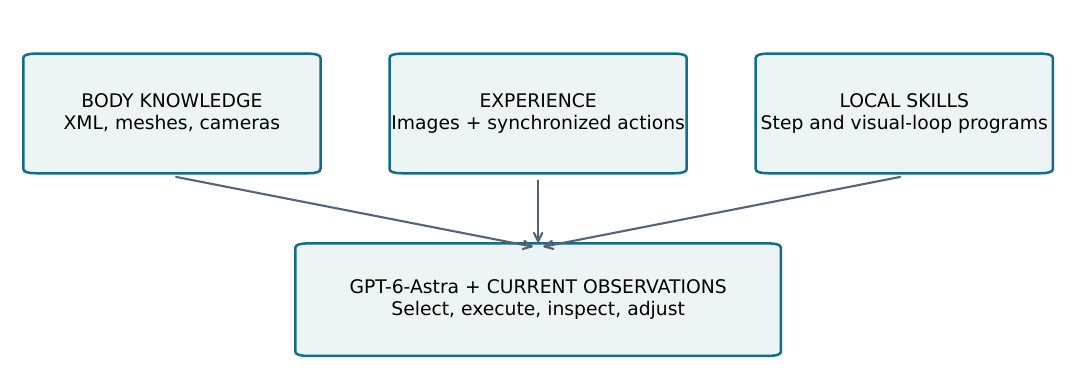}
\caption{Three external resources support execution at different levels: body knowledge informs geometric calculations, recorded experience provides action references, and reusable skills execute local motion and feedback. GPT-6-Astra interprets current observations and selects actions.}
\label{fig:overview}
\end{figure}

We use an elevator-button task that combines mobile approach, arm positioning, visual alignment, and contact. Our platform is a dual-arm XLeRobot with a custom four-wheel mecanum base, a pan--tilt head, and head and wrist cameras. We evaluate it in a MuJoCo elevator lobby and on physical hardware in a different lobby. In both settings, Astra reads a task prompt, inspects the three onboard camera views and joint feedback, and writes programs against a common low-level interface. A local controller executes joint and base commands and returns new observations. Each trial starts in a fresh agent session, with model weights fixed and the supplied materials determined by its experimental condition. Completion time includes reading these materials, model interaction, motion, observation, and outcome confirmation. Figure~\ref{fig:teaser} summarizes the main results; Figure~\ref{fig:overview} outlines the resources available to the agent.

Body information progresses through four levels, I0--I3. I0 provides the common API, actuator names, units, limits, and current feedback. I1 adds a qualitative description of the robot's structure; I2 adds mechanical XML, meshes, and joint mapping; I3 further adds camera geometry and calibration. This progression tests whether a verbal description is sufficient and how much explicit geometry helps Astra compute useful configurations. The full specification remains available as files that the agent can inspect and process with its own code during execution.

Historical experience also progresses through four levels, E0--E3. E0 supplies no prior episode. E1 provides a written summary generated by Astra from a successful attempt. E2 adds sampled images from the three cameras and their visualization, and E3 adds synchronized action commands, robot states, and feedback. The experience representations share one selected source episode, so the comparison examines how its contents are presented for reuse. A combined label such as I0E3 means synchronized experience with no extra body assets. Experience can inform both the initial arm configuration and later decisions about approach and alignment; current observations remain available throughout every trial.

We first compare ten information/experience conditions in 30 simulation trials: I0--I3 without experience, and E1--E3 at the minimum and full body-information levels. We then reuse the same source experience in 18 paired trials at nine new starts displaced by 10--100\,cm. The first comparison measures the value of the supplied resources at a common start; the second examines whether experience remains useful when the approach changes. Complete geometry and synchronized experience yield the largest fixed-start reductions, and the experience-guided agent remains faster at every tested displaced start.

A further question is what Astra can make reusable from its own execution. During a supplementary experience study, it spontaneously writes a routine that alternates short movements with image-based button checks. We refactor this observed routine into STEP, which performs one movement and returns feedback, and LOOP, which repeats movements with an internal visual stopping check. Twenty-seven trials from three near-button states compare these routines against the standard API. This experiment measures the benefit of supplying executable feedback behavior within the same agent-controlled workflow.

Finally, 12 physical trials test sim2real transfer. Conditions A, B, and C respectively provide the common interface alone, simulation body assets, and simulation experience, with three repetitions each at a shared nominal start. Condition D reuses experience from the fastest A trial, once at the original start and once at each of two new starts. The destination task uses an opposite-side DOWN button and operator-confirmed gripper-tip contact, while the source simulation uses an UP button with visual activation feedback. Geometry and experience must therefore be interpreted against the current scene and task. The observed kinematics calculations, historical-image consultation, and subsequent arm and base adjustments explain how GPT-6-Astra uses these resources during transfer.

\begin{samepage}
The study makes three contributions:
\begin{enumerate}
\item A repeated comparison of body knowledge and experience representations, followed by paired tests of experience reuse from new starting positions.
\item A spontaneously generated visual-feedback routine and a controlled evaluation of reusable single-step and repeating-loop skills derived from it.
\item A real-robot study demonstrating sim2real reuse of simulation assets and experience, followed by real-experience transfer to new starting positions.
\end{enumerate}
Together, these experiments identify practical inputs and reusable programs for building manipulation systems around GPT-6-Astra, and connect their timing benefits to the agent's observed use of geometry, recorded experience, and current feedback.
\end{samepage}

\section{Related Work}
Language models can connect task instructions to robot actions through skills, feedback, and generated programs. SayCan~\cite{ahn2022saycan} combines language-model planning with estimates of skill feasibility, while Inner Monologue~\cite{huang2022inner} feeds execution outcomes and scene descriptions back into planning. Code as Policies~\cite{liang2022code} generates robot-control programs containing logic and feedback; VoxPoser~\cite{huang2023voxposer} constructs spatial value maps for motion planning. These systems motivate our program-mediated interface. Within each evaluation block, we hold the agent and low-level interface fixed and vary the robot descriptions, recorded experience, and executable routines supplied to it, measuring how those resources affect execution time and subsequent adjustments.

A complementary line of work incorporates broad knowledge into learned robot models. PaLM-E~\cite{driess2023palme} integrates visual and continuous state inputs into an embodied language model. RT-2~\cite{brohan2023rt2} co-fine-tunes vision--language models on robot actions and web tasks. Open X-Embodiment~\cite{openx2023} pools demonstrations across robots and studies cross-embodiment policy transfer, while Octo~\cite{octo2024} provides a generalist policy that can be adapted to new observation and action spaces. These approaches establish the value of broad training data. We examine a deployment-time question: how much a fixed general-purpose agent benefits when information about its current body and a successful attempt is made available as readable material.

Experience can also be retained outside model weights. ReAct~\cite{yao2023react} interleaves reasoning and interaction with an environment; Reflexion~\cite{shinn2023reflexion} stores verbal reflections for later attempts. Voyager~\cite{wang2023voyager} accumulates an executable skill library in Minecraft. Our experience conditions distinguish a written summary, sampled images, and synchronized image--action records, making representation and access cost part of the comparison. In our study, \emph{emergent skill} refers to a useful local feedback routine that Astra generates during ordinary execution. We parameterize that routine into single-step and repeating-loop forms and test their reuse. Scaling Up and Distilling Down~\cite{ha2023scaling} instead uses language-guided interaction to collect data and distill manipulation policies, connecting generated behavior to subsequent policy learning.

Local visual feedback has a long history in robotics. Classical visual servo control uses image or estimated pose errors to determine robot motion~\cite{chaumette2006visual}. Our STEP and LOOP routines provide a simpler division of work: Astra chooses direction and alignment from images, while the local program executes bounded movement and, in LOOP, checks button activation. This distinction helps interpret the skill results: local stopping feedback saves model interactions when the chosen direction remains useful, while directional revision still requires another agent decision.

Sim2real research addresses differences in appearance, dynamics, and sensing between simulated and physical systems; Zhao et al.~\cite{zhao2020sim2real} survey the principal transfer approaches. Visual domain randomization~\cite{tobin2017domain} and dynamics randomization~\cite{peng2018dynamics} train models across simulation variations to support real deployment. Our transfer uses a different reusable object: explicit robot geometry or multimodal records from one successful simulated attempt. Hardware calibration aligns the joint conventions, and the same agent interprets these materials against fresh real observations with fixed model weights. The simulation and hardware studies therefore examine both what transfers---body configurations and approach references---and what must be adjusted online, including alignment and the completion signal.

\section{System and Experimental Protocol}\label{sec:system}
\subsection{Task, embodiment, and interface}
\begin{figure}[t]
\centering\includegraphics[width=\linewidth]{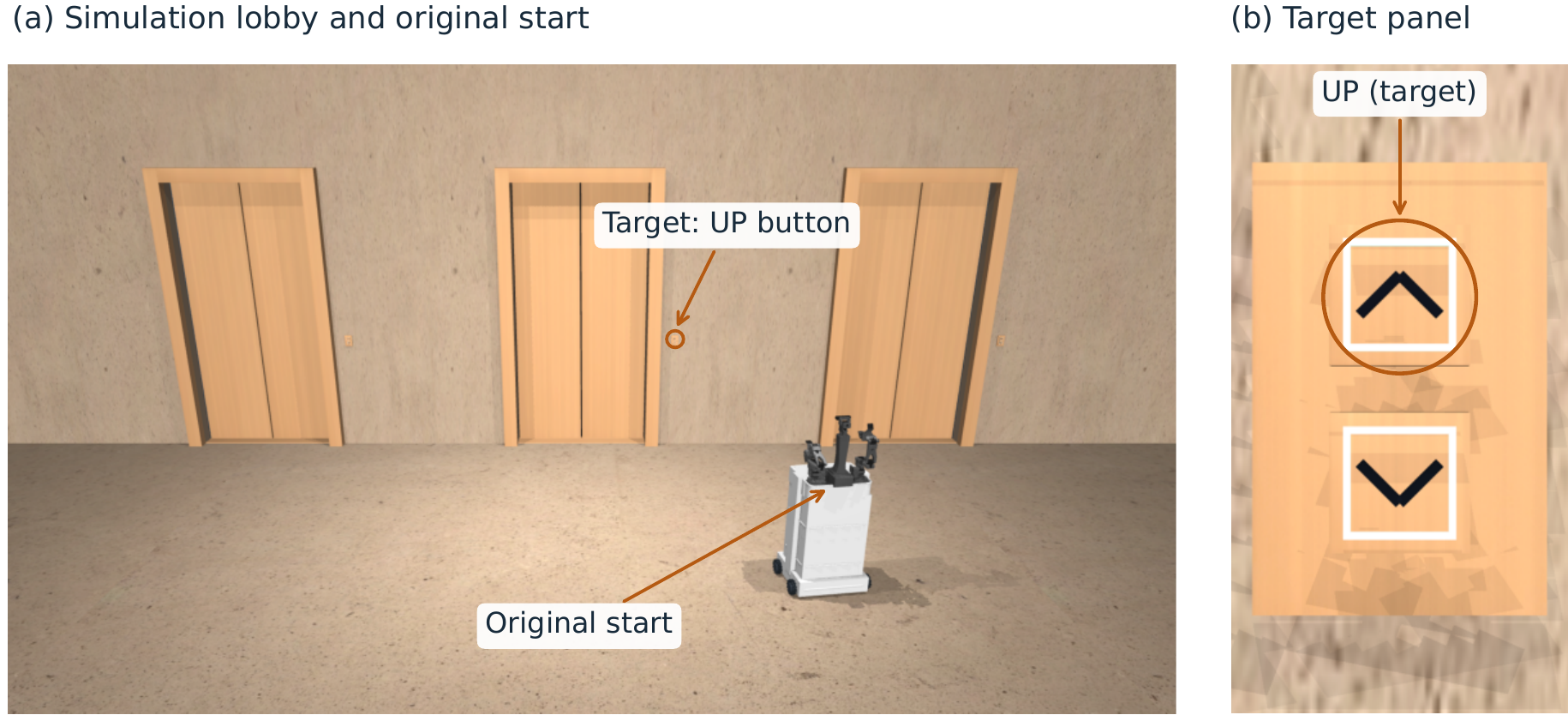}
\caption{The MuJoCo elevator lobby used in the simulation experiments. (a) The cutaway view shows the robot at the original starting state, facing three elevators. (b) The enlarged panel shows the upward call button beside the middle elevator. The agent observes the scene through its head and wrist cameras.}
\label{fig:simulation-scene}
\end{figure}

The simulated and physical platforms are based on XLeRobot~\cite{xlerobot}, with two six-channel arms, a pan--tilt head, and three onboard cameras. We modified the original platform to use a four-wheel mecanum base, enabling forward, lateral, and rotational motion. The MuJoCo~\cite{todorov2012mujoco} environment reproduces the modified robot's geometry and control channels in an elevator lobby (Figure~\ref{fig:simulation-scene}). In both settings, the agent receives head, left-wrist, and right-wrist images on demand, current joint states, and a documented API for timed joint moves, base velocities relative to the robot's current heading, stopping, and task completion. Task observations are limited to this interface; operator monitoring views and evaluator-side simulation state are reserved for supervision and outcome checks.

In simulation, the target is the middle elevator's upward call button. A visible red indicator provides task feedback; evaluator-side activation and wrong-button events provide independent outcome checks. In the real lobby, two elevators are located on each side. The target is the downward call button between the two elevators opposite the initial robot position. Because the physical button is stiff and its light is unreliable, the criterion actually used in the real trials is operator-confirmed contact between the gripper tip and the target button. The task prompt still requests physical depression; Appendix~\ref{app:prompt-real} records this difference between the written instruction and the executed protocol. The operator proactively reports success through the interface; until then, the agent continues adjusting without repeatedly asking for confirmation.

Before hardware evaluation, joint offsets, signs, and ranges are calibrated; wheel geometry and direction are checked; camera roles are confirmed. The camera geometry supplied as prior information comes from simulation. The agent locates the real button from images, without supplied button coordinates. Arms and head are placed in a common task pose before timing starts. A human manually places the base and starts each trial.

\subsection{Body information and experience}
We vary body information (I0--I3) and historical experience (E0--E3), and evaluate their real-robot counterparts in conditions A--D. Table~\ref{tab:design} defines these conditions and summarizes the evaluation blocks. A combined label such as I0E3 specifies both the body-information and experience conditions for a simulation trial.

\begin{table}[t]
\centering\small
\renewcommand{\arraystretch}{1.12}
\begin{tabular}{@{}>{\raggedright\arraybackslash}p{1.3cm}>{\raggedright\arraybackslash}p{\dimexpr\linewidth-1.3cm-2\tabcolsep\relax}@{}}
\toprule
\multicolumn{2}{@{}l}{\textbf{Body information}}\\\midrule
I0 & Common control API, actuator names, units, usable limits, and current feedback.\\
I1 & I0 plus a qualitative description of the robot's structure.\\
I2 & I1 plus mechanical XML, meshes, and joint mapping.\\
I3 & I2 plus camera geometry and calibration.\\\midrule
\multicolumn{2}{@{}l}{\textbf{Historical experience}}\\\midrule
E0 & No historical experience.\\
E1 & Astra-generated written summary of a successful source episode.\\
E2 & E1 plus sampled images from three onboard cameras and their visualization.\\
E3 & E2 plus synchronized action commands, robot states, and feedback.\\\midrule
\multicolumn{2}{@{}l}{\textbf{Real-robot conditions}}\\\midrule
A & Common interface, with no extra body assets or experience.\\
B & A plus simulation body assets corresponding to I3.\\
C & A plus the ordinary simulation E3 package.\\
D & A plus real-robot E3 from the fastest A trial.\\\bottomrule
\end{tabular}

\vspace{0.8em}
\begin{tabular}{@{}>{\raggedright\arraybackslash}p{3.0cm}>{\raggedright\arraybackslash}p{10.3cm}r@{}}
\toprule
Evaluation block & Design & Trials\\\midrule
Fixed simulation start & I0--I3 at E0; E1--E3 at I0 and I3. Three trials per condition. & 30\\
Displaced starts & E0/E3 at nine paired points: three per offset radius. & 18\\
Local skills & FREE/STEP/LOOP at three near-button states, three repetitions each. & 27\\
Real robot & A/B/C: three trials each at a shared start. D1: shared start; D2/D3: two new starts. & 12\\
\bottomrule
\end{tabular}
\caption{Condition definitions and evaluation design. Simulation labels combine a body-information level and an experience level, such as I0E3. The real-robot labels distinguish no prior material, simulation assets, simulation experience, and real experience.}
\label{tab:design}
\end{table}

The written summary is generated by Astra from its own successful execution, closing the loop from acting to summarizing and reusing experience. All simulation experience representations derive from the same source episode, selected under a predeclared rule as the fastest successful episode among three runs at the original starting position (638.352\,s). The source package is then fixed for the controlled comparisons. E2 and E3 include images from 22 sampled times and three onboard views, with each sampled image held until the next sample in a visualization. The same source materials are used for the fixed-start, displaced-start, and simulation-to-real experience comparisons.

For real condition D, the experience source is the fastest of the three A trials. This real experience contains 25 sets of three onboard images and synchronized action and state records, without XML. Each trial starts in a fresh isolated session with a fixed set of supplied materials. The runtime model identifier is \texttt{gpt-6-astra}, with reasoning setting \texttt{xhigh}.

For the local-skill study, FREE supplies no reusable routine, STEP supplies a single short motion followed by feedback, and LOOP supplies repeated motions with an internal visual stopping check. These conditions are evaluated from near-button states in simulation.

\subsection{Comparison design, timing, and evidence}
The fixed-start study uses ten conditions to separate the value of body knowledge from the value of different experience representations. First, I0--I3 are compared at E0 to measure the benefit of progressively richer body information without historical experience. Next, E1--E3 are tested at both I0 and I3 to compare experience representations and determine whether experience remains useful when complete body geometry is already available. The shared I0E0 and I3E0 baselines give four body-information conditions plus six experience conditions, each repeated three times, for 30 trials.

Task time includes material reading, observation, model and service waiting, code generation, actuation, and outcome confirmation. Fixed-start and displaced-start simulation trials use the service task clock, which includes time between actions. Local-skill trials start timing at the agent's first turn. Real trials start at the first formal agent request after pose preparation. Simulation timing ends when the task service accepts the agent's success declaration; real timing ends when the operator's success report is accepted. Preparation, returning the robot to its start, and post-task export are excluded. Time reductions are computed against the corresponding baseline within each evaluation block.

The fixed-start schedule completes the 12 E0 trials, then the nine I0 experience trials, then the nine I3 experience trials. Displaced-start comparisons pair E0 and E3 at each point. Local-skill comparisons rotate condition order across states and repetitions. Real A/B/C use the order ABC, BCA, CAB; D follows selection of the real source.

Simulation success requires target activation, a correct success declaration with fresh evidence, no wrong-button event, and no recorded task fault. Real success uses the gripper-tip contact criterion above, reported online by the operator. We also count motion requests at the API level, treating simultaneous multi-joint targets as one request. Results are reported as means and individual trial times, with sample standard deviations in the supplementary tables. Appendix~\ref{app:prompts} provides the exact task prompts, and Appendix~\ref{app:interface} summarizes the control interface and experience-package contents.

\section{Body Knowledge and Experience in Simulation}
\subsection{Fixed-start information comparison}\label{sec:fixed}
\begin{figure}[t]
\centering\includegraphics[width=\linewidth]{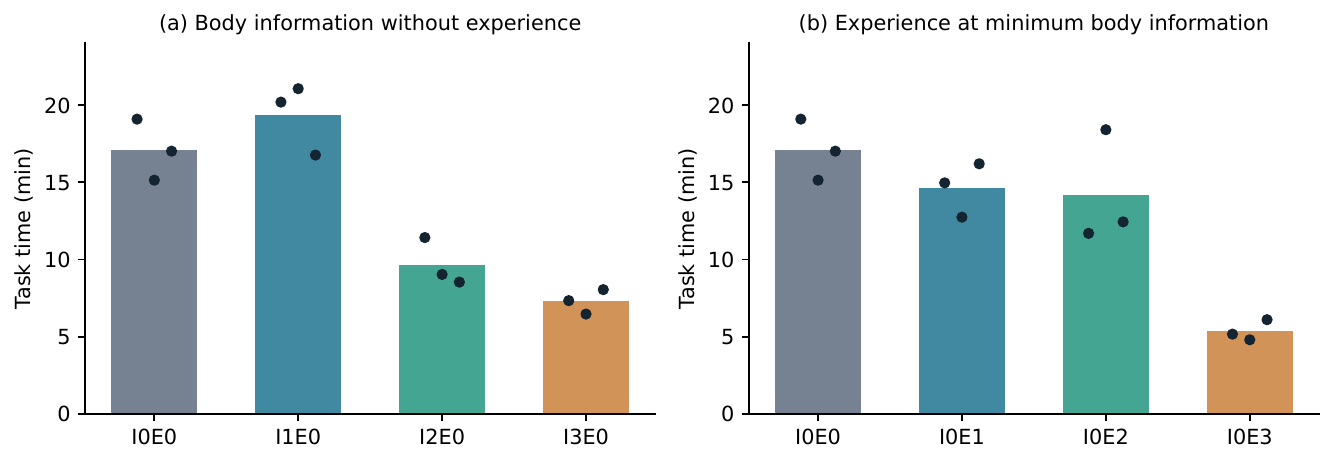}
\caption{Fixed-start simulation results: (a) body-information levels without experience; (b) experience representations at the minimum body-information level. Bars show means; dots show all three trials per condition. Geometry and synchronized action experience have the clearest gains. The full table also includes I3 experience conditions.}
\label{fig:fixed}
\end{figure}

All 30 fixed-start trials succeed. Without historical experience, mean task times for I0, I1, I2, and I3 are 1024.9, 1160.4, 579.8, and 437.1\,s (Figure~\ref{fig:fixed}(a)). Thus mechanical assets alone reduce mean time by 43.4\%, and complete geometry with camera information reduces it by 57.4\% relative to I0. The qualitative I1 condition is not faster, demonstrating that adding information does not automatically reduce execution cost.

At I0, mean times for E0, E1, E2, and E3 are 1024.9, 878.1, 850.7, and 321.4\,s (Figure~\ref{fig:fixed}(b)). E3 reduces mean time by 68.6\% relative to E0. At I3, E3 reduces mean time from 437.1 to 288.9\,s, or 33.9\%. Every E3 observation is faster than every corresponding E0 observation at both body-information levels. In contrast, the smaller differences for written summaries and summaries augmented with sampled images do not show a stable incremental advantage. The complete ten-condition table is in Appendix~\ref{app:fixed}.

\begin{figure}[t]
\centering\includegraphics[width=\linewidth]{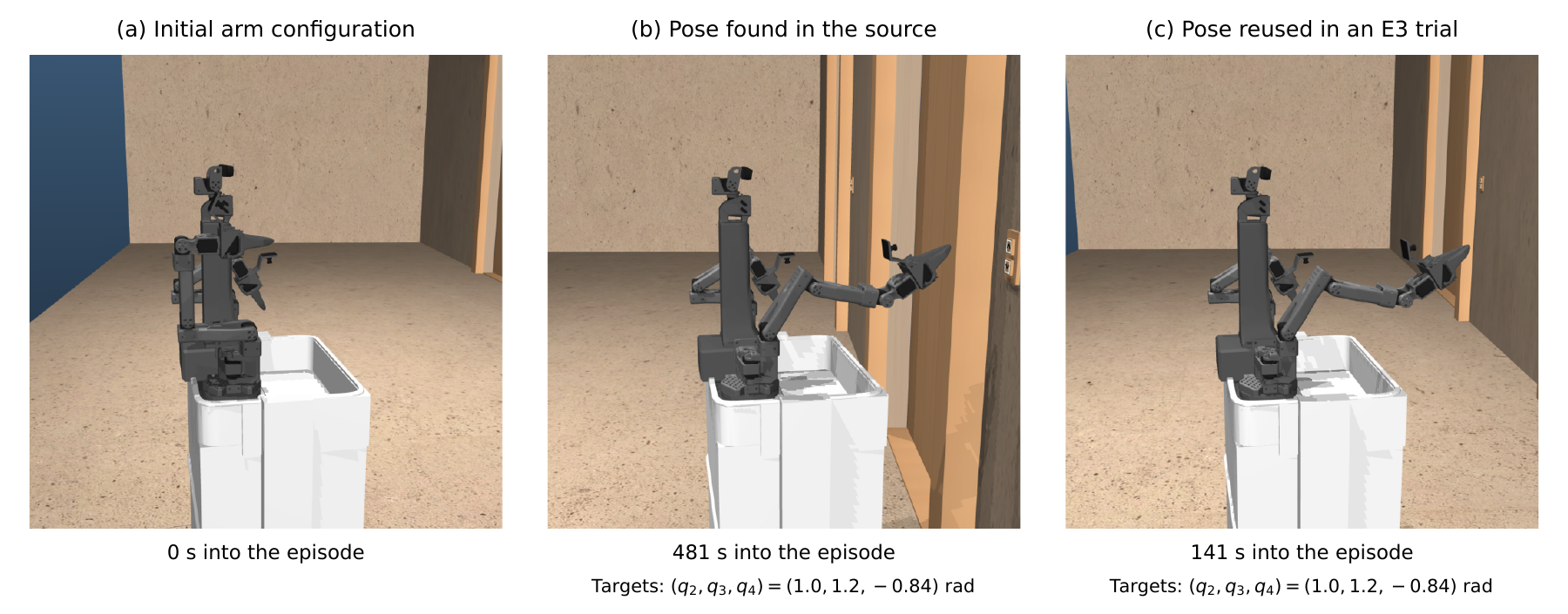}
\caption{\textbf{Reusing a working right-arm pose in simulation.} (a) Initial configuration. (b) The selected experience source at 481\,s, after it has found the arm pose used for pressing. (c) A subsequent I0E3 trial at 141\,s, after adopting the same key joint targets. The shoulder-pitch, elbow, and wrist-pitch targets are $(q_2,q_3,q_4)=(1.0,1.2,-0.84)$ rad in (b,c). Views are rendered from recorded simulation configurations using the same camera angle relative to the robot. The agent continues to adjust the base from current observations.}
\label{fig:arm-pose}
\end{figure}

Command and program records clarify how materials are used. Geometric conditions include XML parsing, coordinate transformations, forward kinematics, and gripper-mesh inspection. All six I0E3/I3E3 trials reuse the successful right-arm target combination $q_2=1.0$, $q_3=1.2$, $q_4=-0.84$ rad, where $q_i$ denotes the angle of joint $i$. The three I0E3 trials each require four joint-move requests, including two head moves, compared with 14 joint actions in the source episode. Figure~\ref{fig:arm-pose}(a)--(c) shows the initial configuration, the working pose found in the source episode, and its early reuse in an I0E3 trial. The recorded configurations make the transfer visible: the right arm is raised and extended toward the panel, with nearly identical shoulder, elbow, and wrist angles in the source and reuse snapshots.

\begin{figure}[tbp]
\centering\includegraphics[width=\linewidth]{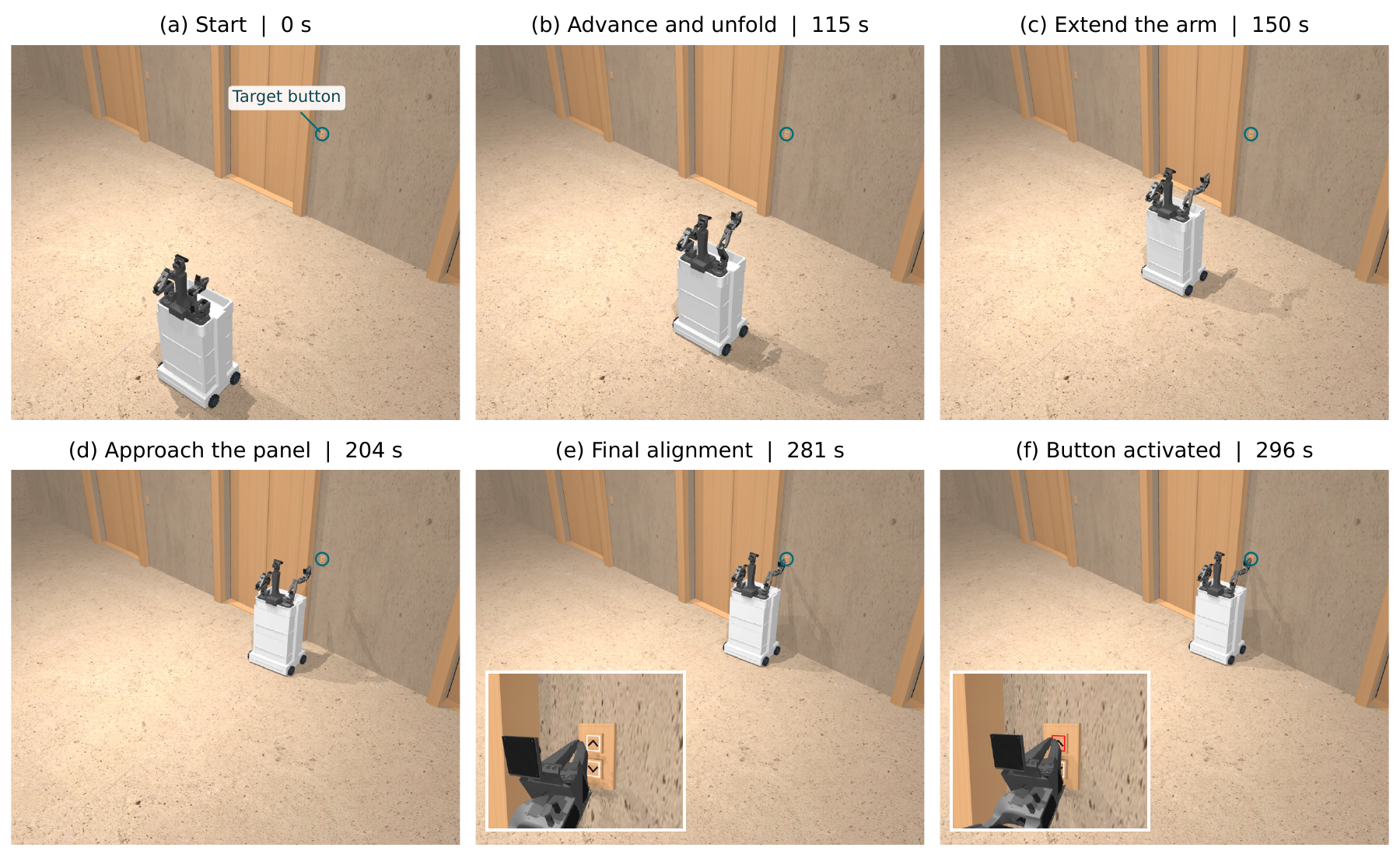}
\caption{A complete simulated approach using recorded experience, completed in 310\,s. (a) Initial position. (b) Advancing with an intermediate arm configuration. (c) Extending into the working pose. (d) Approaching the middle elevator's panel. (e) Final alignment before activation. (f) The gripper activates the upward call button; its indicator turns red. Timestamps use the task clock. Frames are rendered from recorded states with a fixed external camera focused on the robot, middle elevator, and panel; insets in (e,f) enlarge the button interaction.}
\label{fig:simulation-sequence}
\end{figure}

Experience also informs execution after initial pose selection. Figure~\ref{fig:simulation-sequence}(a)--(f) follows the first I0E3 trial from its initial position to button activation. Astra reads historical actions and images, sets an intermediate arm configuration while advancing, and then extends the arm into the working pose (a--c). After approaching the panel and inspecting fresh head and wrist views, it reopens historical head and right-wrist images before issuing further short advances and lateral corrections (d,e). Near contact, separately observed movements become shorter, and the agent continues past apparent gripper--button overlap until current images show activation (f). Historical images thus serve as references during approach and contact preparation, alongside the synchronized action record.

The six fixed-start E3 trials all use two arm-deployment stages with a fresh observation between them; five also issue six base commands between the stages. Subsequent alignment remains responsive to current images. This pattern combines a reusable deployment sequence with online approach decisions. Task time also depends on work between motion requests: I3E3 finishes sooner than I3E0 despite making more requests on average (37.3 versus 33.3), reflecting differences in observation, model interaction, and service communication as well as actuation.

\subsection{Reuse at displaced starts}\label{sec:displaced}
To test whether experience remains useful away from its source position, we reuse the original-start E3 package at nine new starts. Three points are selected in separated directions on each of the 10, 50, and 100\,cm circles around the original base position. Initial heading and arm posture are held fixed. At each point, one trial receives E3 and one receives no experience, with body information held at I0. This produces nine matched comparisons and 18 trials; directions are sampled separately for each radius.

\begin{figure}[t]
\centering\includegraphics[width=\linewidth]{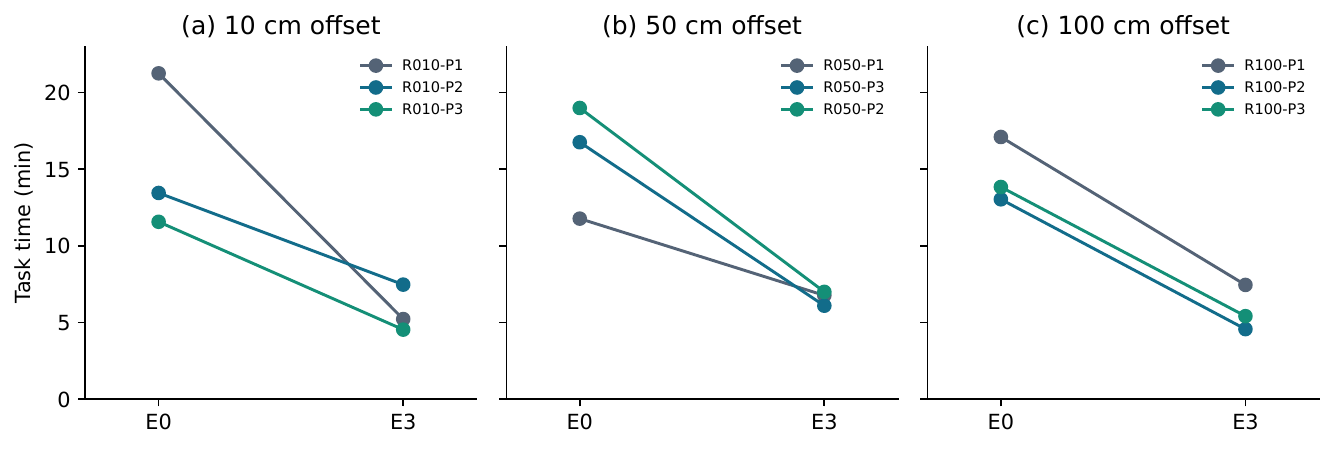}
\caption{Experience reuse from new starting positions at (a) 10\,cm, (b) 50\,cm, and (c) 100\,cm offset radii. Each line compares a trial without experience (E0) with a trial using experience recorded at the original start (E3), from the same displaced position. Colors identify points within each offset radius. Experience reduces task time at every tested point.}
\label{fig:perturbation}
\end{figure}

All 18 trials succeed. Mean times without/with original-start experience are approximately 15:25/5:44 at 10\,cm (Figure~\ref{fig:perturbation}(a)), 15:50/6:37 at 50\,cm (Figure~\ref{fig:perturbation}(b)), and 14:39/5:48 at 100\,cm (Figure~\ref{fig:perturbation}(c)). Experience is faster at all nine points, reducing mean time by 58--63\% across the three offset groups. Thus the recorded experience generalizes to the tested displaced starts while the agent adjusts its approach from current images.

The staged deployment also persists across starts. All nine displaced-start E3 trials use an intermediate arm configuration, observe again, and then reach the key right-arm combination in Figure~\ref{fig:arm-pose}(b,c). Across the six fixed-start and nine displaced-start E3 trials, 11 of 15 include six to ten base commands between these stages. Once deployed, the arm targets stay fixed in these simulation trials while the base performs the remaining alignment. Average joint requests at displaced starts fall from 24.67 to 4.56 (81.5\%), and base requests from 36.22 to 31.89 (12.0\%). The largest reduction in motion requests therefore comes from avoiding repeated arm search, within a broader process that uses historical visual references and current observations to organize the approach. Separating the benefit of this process from that of the working pose alone requires the ablations discussed in Section~\ref{sec:limitations}.

\section{From Emergent Behavior to Reusable Skills}
\subsection{A spontaneously generated local loop}
During a supplementary simulation study comparing experience packages (Appendix~\ref{app:reflection}), Astra writes a program that advances briefly, captures images, checks red pixels within selected regions for button activation, and repeats until a threshold is reached or the step budget is exhausted. The program permits at most five iterations; the observed episode executes four, taking about 4.6\,s locally, followed by Astra's visual confirmation. The agent has combined motion, observation, and a perceptual stopping rule into a local feedback routine during task execution.

This local routine can avoid intermediate model round trips. If $L$ is the additional latency of each model-mediated decision, four equivalent steps with three intermediate decisions give
\begin{equation}
T_{\mathrm{stepwise}}\approx T_{\mathrm{local}}+3L,
\qquad \Delta T\approx3L.
\end{equation}
Using 15--25\,s inter-call intervals observed in later records gives an estimated 45--75\,s reduction in intermediate decision latency under this four-step scenario. This calculation covers the three avoided model round trips; total task time also includes initial observation, program construction, and final confirmation. The next experiment measures end-to-end performance when such routines are made available for reuse.

\subsection{Evaluating reusable skills}\label{sec:skills}
We refactor the observed behavior into parameterized routines with explicit motion and time limits, then compare three conditions. FREE gives Astra the standard API and freedom to write control code, including its own loops. STEP supplies a routine that executes one short base movement, checks feedback, and returns a new image. LOOP uses the same movement routine and repeats it internally with a red-pixel stopping test. Both routines limit translation to 0.06\,m/s and each pulse to 0.1--0.8\,s. Astra selects the direction, speed, and duration; for LOOP it also selects the image region and execution budget. Astra confirms the final outcome in both cases.

Three near-button states, S1--S3, are derived from a recorded trajectory distinct from the discovery episode. The base is shifted backward by 0, 2, and 4\,cm, respectively, while retaining a common arm pose, so the states test progressively longer final approaches. Simulator state is restored and verified before each trial. Each state--condition combination is repeated three times, for 27 trials. Timing covers reading the supplied materials, selecting parameters, executing actions, and confirming success. Table~\ref{tab:local-states} reports the mean and sample standard deviation for each state and condition.

\begin{figure}[t]
\centering\includegraphics[width=.92\linewidth]{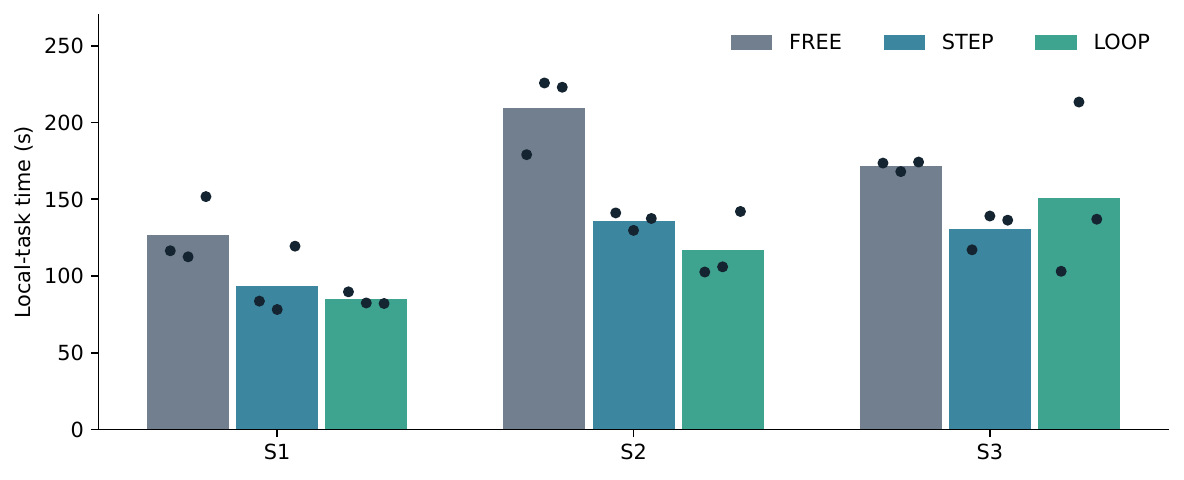}
\caption{Near-button task times with no supplied routine (FREE), a single-step skill (STEP), or an internally repeating skill (LOOP). S1--S3 progressively increase the initial distance to the button; each state--condition combination has three trials. STEP and LOOP reduce the overall mean, with most of the gain already achieved by STEP.}
\label{fig:skills}
\end{figure}

All 27 trials succeed. Across states, FREE, STEP, and LOOP average 169.4, 120.2, and 117.6\,s (Figure~\ref{fig:skills}). STEP reduces mean time by 29.0\% and is faster in all nine matched state--repetition comparisons; LOOP reduces it by 30.6\% and is faster in eight of nine. Most of the mean improvement is already achieved by STEP: LOOP's mean time is 2.2\% lower than STEP's, and it wins four of their nine comparisons.

At S3, LOOP averages 151.1\,s versus STEP's 130.8\,s, mainly because one trial takes 213.3\,s and requires seven calls to revise direction and alignment (Appendix~\ref{app:skill-diagnostics}). LOOP checks when to stop but holds direction fixed within each call, so these repeated returns to Astra offset its savings in model round trips.

\section{Sim2Real and Real-Experience Transfer}
\subsection{Real-lobby protocol}\label{sec:real-protocol}
We evaluate the physical XLeRobot in a real elevator lobby after the calibration and interface preparation described in Section~\ref{sec:system}. Hardware calibration expresses joint targets in the same XML reference as simulation, and both APIs use body-frame base velocities. The task changes from the middle elevator's UP button to the opposite-side DOWN button, with success determined by operator-confirmed gripper-tip contact. The controller allows velocity updates during continuous base motion and stops when a command expires or a stop is requested. The translation speed cap is 0.10\,m/s for all 12 trials.

The three repetitions of conditions A, B, and C compare the baseline, simulation body assets, and simulation experience from the same nominal base start, with manual placement variation. For condition D, the fastest baseline trial supplies real experience. D1 starts at the shared location. D2 and D3 start at two other positions selected to test reuse across placements. These new positions are recorded qualitatively; their coordinates and headings were not surveyed.

\begin{table}[t]
\centering
\begin{tabular}{llrrrr}
\toprule
Condition & Material & Trial 1 & Trial 2 & Trial 3 & Mean\\\midrule
A & None & 808.9 & 431.5 & 980.1 & 740.1\\
B & Simulation XML & 284.3 & 408.5 & 351.2 & 348.0\\
C & Simulation E3 & 366.5 & 381.3 & 364.3 & 370.7\\
D & Real E3 & 250.3 & 269.7 & 360.0 & $293.4^{a}$\\\bottomrule
\end{tabular}
\par\vspace{3pt}
\begin{minipage}{.88\linewidth}\footnotesize
\textsuperscript{a}D's mean summarizes three placements: D1 shares the A/B/C start; D2 and D3 use two new starts. It is a descriptive mean across placements.
\end{minipage}
\caption{Real-task times in seconds, using operator-confirmed gripper-tip contact as success. A/B/C each have three repeats at a shared nominal start.}
\label{tab:real}
\end{table}

\subsection{Same-start efficiency and material use}\label{sec:real-use}
\begin{figure}[t]
\centering\includegraphics[width=\linewidth]{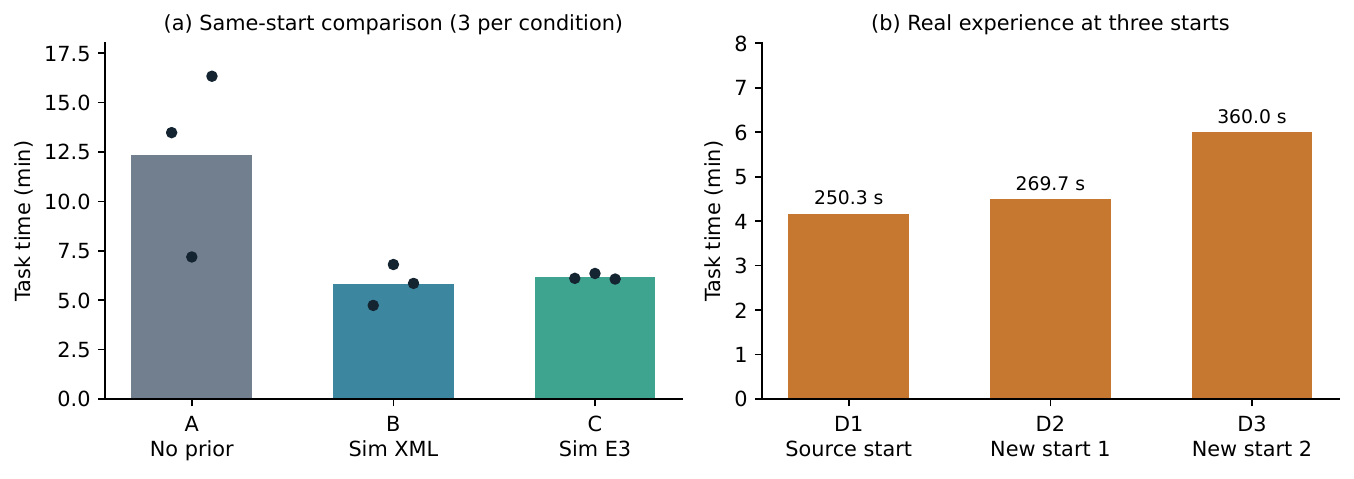}
\caption{Real-robot results. (a) Baseline, simulation assets, and simulation experience at a shared nominal start, with three trials per condition. (b) Real experience reused once at the source start and once at each of two new starts. D2/D3 test transfer across placements.}
\label{fig:real}
\end{figure}

All 12 trials reach operator-confirmed contact with the target button (Table~\ref{tab:real}). In the same-start comparison, the baseline, simulation-asset, and simulation-experience conditions average 740.1, 348.0, and 370.7\,s (Figure~\ref{fig:real}(a)). Simulation assets and simulation experience therefore reduce mean time by 53.0\% and 49.9\%, respectively. Every asset-guided and experience-guided trial is faster than the fastest baseline trial. The three simulation-experience times span 17.0\,s.

All three asset-guided trials parse the supplied XML and write forward-kinematics calculations; the latter two also inspect gripper mesh geometry. These programs convert the simulation's body hierarchy and joint transforms into estimates of physical arm configurations under the calibrated joint convention. Average arm/head move requests fall from 20.0 in the baseline to 3.7 with assets, and base requests from 18.7 to 10.7. The recorded calculations and reduced adjustment counts link the transferred body description to finding a useful arm configuration at the real button.

All three simulation-experience trials read the source action history. The first two each view a late source head image, and the third views three source head images. Their commands expose both reuse and adaptation. In the second trial, Astra adopts the simulation targets $(q_2,q_3,q_4)=(1.0,1.2,-0.84)$ rad at about 104\,s, then changes them to $(1.65,1.2,-0.19)$ rad at about 143\,s while adjusting its approach from current images. The first two trials also move left initially and subsequently correct to the right, consistent with carrying over a source-related directional preference that requires correction in the new lobby. This detour illustrates that experience can introduce unhelpful approach choices as well as useful configurations (Section~\ref{sec:limitations}). Live observations guide subsequent changes to the arm and base. All three complete the current DOWN-button task under operator confirmation despite receiving experience of the simulated UP-button task.

\subsection{Real experience beyond its source start}\label{sec:real-reuse}
D1 completes in 250.3\,s at the original start, while D2 and D3 complete in 269.7 and 360.0\,s at their different new starts (Figure~\ref{fig:real}(b)). All three read real source records and historical images. D2 directly adopts a source arm combination $(q_2,q_3,q_4)=(2.07,1.9,-0.11)$ rad; D1 also reaches that combination later. D3 uses different intermediate targets and continues adjusting. These executions show that experience recorded at one physical starting position can support successful approach and button contact from other positions in the lobby.

\begin{figure}[t]
\centering\includegraphics[width=.85\linewidth]{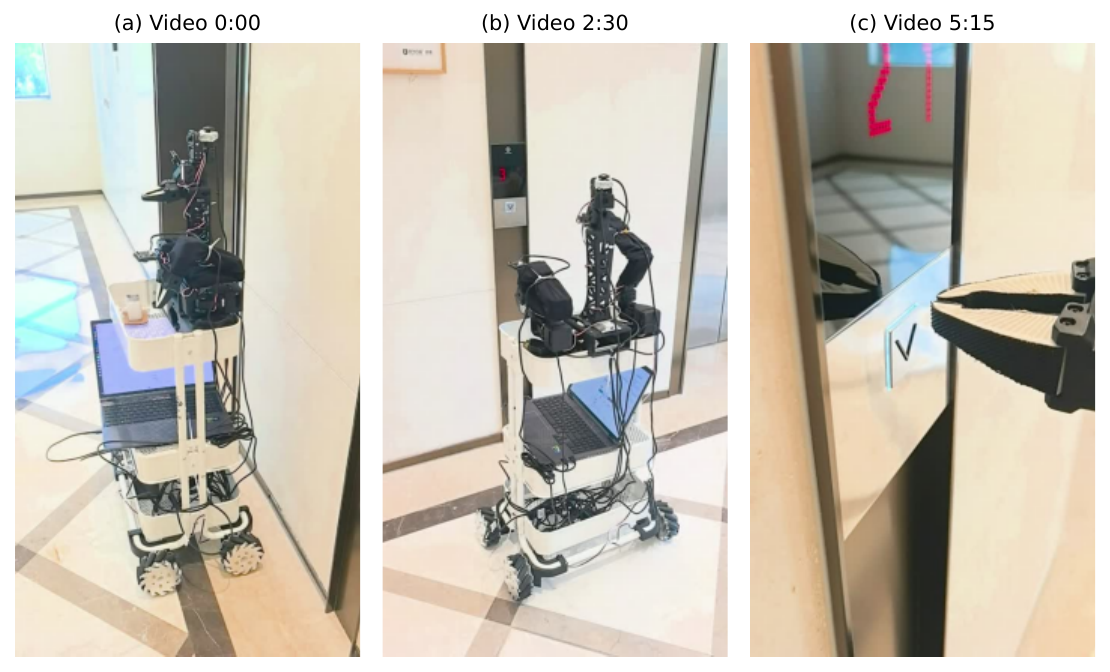}
\caption{External views of real-experience transfer in D3: (a) initial view, (b) approach, and (c) near-button interaction. Labels indicate time within the demonstration video.}
\label{fig:demo}
\end{figure}

The release includes an external video of D3 at original speed with audio removed and a silent six-times-speed version. Figure~\ref{fig:demo}(a)--(c) shows three stages of this approximately 336\,s recording, illustrating the approach and button interaction; the reported 360\,s task time comes from the experiment timestamps. Three onboard camera recordings and timestamp records are retained in the research archive, and the publication package provides the selected demonstration and action records.

\section{Discussion}
\paragraph{Why sim2real reuse works in this task.}
The real-robot reductions of 53.0\% with simulation assets and 49.9\% with simulation experience show that both resources remain useful across the change of lobby and task feedback. The behavior in Sections~\ref{sec:fixed} and~\ref{sec:real-use} suggests three reasons for GPT-6-Astra's strong sim2real performance here: it interprets explicit robot descriptions, extracts configurations and approach references from multimodal records, and revises actions through current observations. XML parsing and generated kinematics code connect body knowledge to motion; historical image and action reads support both deployment and later visual comparison. These operations make simulation materials usable during physical execution with fixed model weights. In the domain-randomization studies~\cite{tobin2017domain,peng2018dynamics}, simulation variation shapes a trained model before deployment. Here, the reusable descriptions and records remain available during execution, allowing Astra to reinterpret them as the observed geometry and task feedback change.

\paragraph{Transferring body knowledge and working configurations.}
Robot geometry and joint configurations retain their meaning across the calibrated simulation and hardware interfaces (Section~\ref{sec:real-protocol}). The simulation results identify useful units of reuse: E3 supplies a staged deployment culminating in the working pose (Figure~\ref{fig:arm-pose}), and joint requests decrease much more than base requests at displaced starts (Section~\ref{sec:displaced}). Historical images remain available as visual references during the approach. The real trials show the corresponding behaviors: assets support new geometric calculations, while experience supplies an arm configuration that Astra subsequently adjusts. Reusing body-relative information reduces dependence on reproducing the source starting position or full motion sequence.

\paragraph{Grounding transferred information in the current task.}
The simulation-experience condition on hardware combines an UP-button demonstration with the current DOWN-button instruction and operator feedback. Changed shoulder and wrist targets show active adaptation in the new lobby (Section~\ref{sec:real-use}); the left-then-right corrections also expose the cost of carrying over an approach choice that needs revision. The two new real starts extend the adaptation pattern: a source pose is reused in one trial, while another uses different intermediate targets (Section~\ref{sec:real-reuse}). These observations support an interpretation in which Astra selects reusable parts of the source and grounds subsequent decisions in the current scene. Transfer can preserve useful body knowledge while changing alignment, approach, and the signal used to confirm completion.

\paragraph{Reusable programs within the same control design.}
The skill study (Section~\ref{sec:skills}) reveals a related division of work. STEP captures most of the mean improvement. LOOP can remove intermediate decisions when its parameters remain appropriate, but its success detector does not provide directional feedback; repeated alignment corrections explain the slower S3 mean. A more capable routine would estimate progress and alignment, adjust direction locally, or return early when its assumptions fail. Classical visual servo control~\cite{chaumette2006visual} offers a relevant design principle: measured alignment error can guide the next movement as well as determine when to stop. Such a revision, together with completion checks selected for the destination environment, is a concrete next step toward transferring local skills to hardware.

\section{Limitations}\label{sec:limitations}
The study covers one robot platform and task family, with three repetitions per main condition and one selected experience source per domain. Experience can also carry over inefficient choices, as illustrated by the left-then-right corrections in two real trials. Pose-only and preset-pose controls would help separate the benefit of a useful arm configuration from that of richer experience during approach. The two new real starts have no matched baseline, and hardware success uses operator-confirmed contact rather than verified button depression. Broader tests across sources, headings, and embodiments, together with local alignment feedback for LOOP, would extend the present findings.

\section{Conclusion}
Robot knowledge, successful experience, and reusable skills reduce repeated exploration in manipulation controlled by GPT-6-Astra. In simulation, complete body information and synchronized experience reduce mean task time by 57.4\% and 68.6\%, respectively; experience recorded at the original start also yields 58--63\% reductions from the tested displaced starts. Reusable skills derived from an agent-generated routine reduce near-button task time by 29--31\%. On the physical robot, simulation assets and experience reduce same-start mean time to operator-confirmed button contact by 53.0\% and 49.9\%, demonstrating sim2real transfer, and real experience supports execution from two new starts. For general-purpose manipulation experiments built around GPT-6-Astra, the results favor providing explicit body files and synchronized successful episodes, then packaging useful agent-generated feedback routines for reuse. This gives the agent concrete material to inspect, calculate with, and adapt as current observations change.

\begingroup
\small
\setlength{\parskip}{0pt}
\setlength{\bibsep}{4pt}
\interlinepenalty=10000
\bibliographystyle{unsrtnat}
\bibliography{references}
\endgroup
\clearpage
\appendix
\small
\setlength{\parskip}{6pt plus 1pt minus 1pt}
\section{Complete Fixed-Start Results}\label{app:fixed}
Table~\ref{tab:full} reports all ten tested conditions. At I3, the written-summary and sampled-image conditions have higher mean times than the no-experience condition, while synchronized image--action experience is faster. This pattern motivates distinguishing experience representations: the tested summaries, sparse images, and synchronized records differ in both their content and reading cost.
\begin{table}[H]\centering\small\begin{tabular}{lrrrr}\toprule
Condition & $n$ & Mean (s) & SD (s) & Motion requests \\ \midrule
I0E0 & 3 & 1024.9 & 118.7 & 66.7 \\
I1E0 & 3 & 1160.4 & 136.3 & 67.3 \\
I2E0 & 3 & 579.8 & 92.5 & 48.0 \\
I3E0 & 3 & 437.1 & 47.5 & 33.3 \\
I0E1 & 3 & 878.1 & 105.2 & 55.0 \\
I0E2 & 3 & 850.7 & 220.7 & 57.7 \\
I0E3 & 3 & 321.4 & 40.2 & 35.0 \\
I3E1 & 3 & 541.7 & 138.6 & 41.0 \\
I3E2 & 3 & 499.8 & 30.8 & 35.7 \\
I3E3 & 3 & 288.9 & 41.9 & 37.3 \\
\bottomrule\end{tabular}

\caption{Complete fixed-start results. SD is sample standard deviation. Every condition succeeds 3/3; time uses the simulation service task clock.}\label{tab:full}\end{table}

\section{Local States and Skill Semantics}\label{app:skill-diagnostics}
The discovery routine and evaluation states come from two different trials in the supplementary experience-compression and reflection study (Appendix~\ref{app:reflection}). In a compact-experience trial from the first measurement, Astra spontaneously generated the local visual-feedback routine. A separate compact-experience trial from the second measurement supplies a near-button snapshot at 223.15\,s; backward base offsets of 0, 2, and 4\,cm from that snapshot produce S1--S3. For archive lookup, these are \texttt{er001-t005} (routine discovery) and \texttt{er001-t016} (state source). The identifiers label recorded trials, not additional experimental conditions. The simulator restores positions, velocities, control state, and button appearance, then verifies that the button is initially inactive. An offline check establishes local reachability, while Astra selects the direction and control parameters during evaluation. Each local trial has a 300\,s task budget. Full approach tasks use a 1800\,s budget in the main information and real studies.

STEP and LOOP accept translational speed norm at most 0.06\,m/s and pulse duration 0.1--0.8\,s. The shared implementation enforces execution limits and stopping. LOOP detects the red indicator using pixels satisfying $r>160$, $r>1.7g$, and $r>1.7b$, with more than 30 qualifying pixels in the agent-selected region. It checks the initial image and enforces step and time budgets. Astra selects the image region containing the target button; the local routine checks its color feedback after each motion pulse.

\begin{table}[H]\centering
\begin{tabular}{lrrr}\toprule
State & FREE & STEP & LOOP\\\midrule
S1 & $126.9\pm21.6$ & $93.8\pm22.4$ & $84.8\pm4.3$\\
S2 & $209.2\pm26.2$ & $136.1\pm5.8$ & $116.9\pm21.9$\\
S3 & $171.9\pm3.4$ & $130.8\pm12.0$ & $151.1\pm56.5$\\\bottomrule
\end{tabular}
\caption{Local-task mean and sample SD in seconds, three trials per cell.}\label{tab:local-states}
\end{table}

\begin{table}[H]\centering\small
\begin{tabular}{lrrrr}\toprule
S3 repetition & STEP (s) & LOOP (s) & LOOP calls & Motion pulses\\\midrule
1 & 117.1 & 103.1 & 3 & 5\\
2 & 139.1 & 213.3 & 7 & 9\\
3 & 136.4 & 137.0 & 5 & 13\\\bottomrule
\end{tabular}
\caption{S3 diagnostic breakdown. Calls count separate invocations by Astra; pulses count movements inside those invocations. The second repetition drives most of LOOP's larger S3 mean.}\label{tab:s3-diagnostics}
\end{table}

In S3 repetition 2 (Table~\ref{tab:s3-diagnostics}), the seven calls select backward, forward, forward, right, left, right, and right motion, executing 1, 1, 2, 1, 1, 1, and 2 pulses, respectively. The first six calls end at their step budgets; the last detects activation. The image region is also revised between calls. Summing simulator-clock intervals from each call's first base request to its returned observation gives 9.54\,s; the intervening gaps sum to 130.31\,s. These intervals separate local execution from the much longer work between calls, without attributing all of the latter to model computation. The released diagnostic records contain the parameters, timestamps, activation-to-detection delays, and the staged-deployment and historical-image evidence used in Section~\ref{sec:fixed}.

\section{Exploratory Experience Comparisons}\label{app:exploratory}
\subsection{Earlier iterative written experience}
An earlier six-episode simulation sequence accumulated written experience between sessions. The first five episodes succeeded in 18:37, 16:04, 14:14, 14:10, and 19:54; the sixth ended in declared failure at 27:36. Repeating the input versions used for the fourth and fifth episodes three times each produced all successes with means 16:29 and 17:19, a small difference relative to trial variation. Accumulating written experience therefore produced mixed results, motivating the controlled representation comparisons in the main study. These exploratory results are reported separately from the primary evaluation blocks.

\subsection{Initial target information}
An exploratory four-condition test supplied body information, the button's initial pose, both, or neither, with one trial per condition. Times were 11:15 without either, 7:10 with body information, 15:21 with the initial button pose, and 5:26 with both; all succeeded. Body geometry and supplied target pose can support overlapping calculations, since geometry and current images also allow online relative-pose estimation. The fastest observed execution used both forms of information.

\subsection{Experience compression and reflection}\label{app:reflection}
A supplementary comparison used ordinary E3, a compact experience package, and an expanded reflection guide at six test points. The first measurement reused six historical E3 results and added 12 compact/reflection trials; a second measurement ran 18 interleaved E3/compact/reflection trials at the same points. All new trials succeeded. Times use the agent execution clock, from the first turn to accepted task completion. First-measurement means were 370.35, 384.53, and 404.30\,s; second-measurement means were 354.47, 325.63, and 413.67\,s. Compact experience was faster than expanded reflection in 10/12 pointwise comparisons, and faster than ordinary E3 at 3/6 points in each measurement. The packages change both content and organization, including selected images and reading cost; ordinary E3 already contains qualitative guidance. This study also supplied the episode in which Astra generated the local feedback routine subsequently evaluated as a reusable skill.

\section{Task Prompts}\label{app:prompts}
The following are the complete task prompts preserved with the experimental inputs. Wording is unchanged; only typesetting is adapted. Each prompt is accompanied by the API documentation and a condition-specific list of authorized materials.
\par\noindent\begin{minipage}{\linewidth}
\subsection{Simulation approach and pressing}\label{app:prompt-sim}
The fixed-start and displaced-start studies use this common task prompt. The condition-specific \texttt{PRIOR.md} lists the authorized materials defined in Table~\ref{tab:design}.
\begin{list}{}{\leftmargin=1em\rightmargin=1em}\item[]\small
Physically control XLeRobot to approach the middle elevator and press only its UP call button. Verify from fresh observations that the target button has turned red before declaring success. Work alone without questions; minimize time and online decisions without damage.

Read API.md and PRIOR.md. Use the shared low-level interface, explicitly supplied information, installed dependencies and code you write. Choose your observation and action strategy freely; no exploration or model-building phase is required. Keep head pan within $\pm$90 degrees of its fresh initial orientation and use the permitted tilt range.

Only the historical material explicitly listed in PRIOR.md is authorized. No other conversations, workspaces, skills, prior Git history, external controllers/information, unprovided simulation internals/assets, datasets, demonstrations, pretrained policies or hidden evaluation. Historical observations are not current observations; verify action results through current feedback. Stop on contamination or unavailable essential capabilities. Never bypass protections or reset; the whole session is one attempt.

On termination, stop safely and call finish. Then save a brief report in evidence/ describing outcome, fresh visual verification, final state, elapsed time, motion commands, decision points and which supplied information was used. Do not produce a skill for another attempt.
\end{list}
\end{minipage}\par\addvspace{1em}

\par\noindent\begin{minipage}{\linewidth}
\subsection{Local-skill evaluation}\label{app:prompt-local}
FREE, STEP, and LOOP use identical task prompts. In STEP and LOOP, \texttt{PRIOR.md} additionally points to \path{skill/local-press/SKILL.md}; the agent can read the supplied routine or write its own code.
\begin{list}{}{\leftmargin=1em\rightmargin=1em}\item[]\small
Complete only the final pressing stage for the middle elevator UP button. The robot is already near the correct panel with its right gripper in a working pose; no row-identification tour or arm deployment is required. Use fresh images to judge alignment and choose bounded motion. Only the UP button may be activated. In this simulation activation is indicated by a red border. Verify a fresh image, stop, then finish(success). Minimize task time and online decisions without damage.

Read API.md and PRIOR.md. Work alone. You may freely write programs, including image-processing feedback loops. Keep head pan within $\pm$90 degrees of its fresh initial orientation. All groups have the same public interface and joint position/limit feedback, with no geometry model or target coordinates. Do not assume commanded velocity times duration equals actual displacement.

Use only the provided interface, installed dependencies, explicitly supplied skill if any, and code written in this workspace. Do not access other experiments, conversations, Git history, hidden state/assets, external information or controllers. Never reset or bypass protections. Keep supplied files unchanged. Stop on unavailable essential capabilities or contamination. At termination stop safely, call finish, and write a brief evidence/report.md with images, actions, decisions and outcome. Do not create a transfer artifact for another trial.
\end{list}
\end{minipage}\par\addvspace{1em}

\par\noindent\begin{minipage}{\linewidth}
\subsection{Real-robot evaluation}\label{app:prompt-real}
Conditions A--D share this task prompt; their authorized materials differ as in Table~\ref{tab:design}. The prompt below requests physical button depression. In the executed hardware trials, the button's stiffness led the operator to accept gripper-tip contact with the target button as success. The reported real-robot results use that contact criterion; the original prompt is reproduced unchanged.
\begin{list}{}{\leftmargin=1em\rightmargin=1em}\item[]\small
Physically control XLeRobot to press the DOWN call button between the two elevators on the side opposite its initial position, not the button on its starting side. Continue if the button is already lit or people pass through. Minimize time and online decisions without damage.

The operator proactively reports success through API.md. Success requires physically depressing the target button; visual overlap or initial contact alone is insufficient. Unless the operator reports success, treat the press as unsuccessful and continue adjusting without requesting confirmation. On success, a wrong-button report or a stop request, stop safely and end the attempt.

Read API.md and PRIOR.md. Use the shared low-level interface, explicitly supplied information, installed dependencies and code you write. Choose your observation and action strategy freely; no exploration or model-building phase is required. Keep head pan within $\pm$90 degrees of its fresh initial orientation and within the real interface limits; use the permitted tilt range. Decide and act independently without requesting hints; operator outcome and stop events are authorized feedback.

Only the historical material explicitly listed in PRIOR.md is authorized. No other conversations, workspaces, skills, prior Git history, external controllers/information, unprovided simulation internals/assets, datasets, demonstrations, pretrained policies or hidden evaluation. Supplied models and historical experience retain the provenance stated in PRIOR.md; camera calibration applies only to its documented configuration. The current task definition, real API, current observations and current operator outcome events take precedence over historical scene descriptions and action values. Historical observations are not current observations; verify action results through current feedback. Stop on contamination or unavailable essential capabilities. Never bypass protections or reset; the whole session is one attempt.

On termination, stop safely and end the attempt using the procedure in API.md. Then save a brief report in evidence/ describing outcome, operator confirmation event and its timestamp, final state, elapsed time, motion commands, decision points and which supplied information was used. Do not produce a skill for another attempt.
\end{list}
\end{minipage}\par\addvspace{1em}

\section{Interface and Experience-Package Details}\label{app:interface}
\subsection{Control interface}
Table~\ref{tab:api} summarizes the operations common to the simulation and real interfaces. Joint angles use a fixed XML reference, rather than the robot's startup pose. Base velocities use the robot's current frame: positive $v_x$ is forward, positive $v_y$ is left, and positive $\omega_z$ is counterclockwise. The agent receives joint feedback and commanded base velocity, without global base odometry or target coordinates.

\begin{table}[H]\centering\small
\begin{tabular}{@{}>{\ttfamily}p{1.6cm}>{\raggedright\arraybackslash}p{3.6cm}>{\raggedright\arraybackslash}p{9.0cm}@{}}
\toprule
\normalfont Operation & Arguments & Returned information or effect\\\midrule
schema & None & Camera roles, joint bounds, and motion limits.\\
state & None & Timestamped joint positions, targets, and motion status.\\
observe & None & Fresh head, left-wrist, and right-wrist images with capture timestamps.\\
move & Joint targets; duration & Coordinated bounded joint motion; returns before execution finishes.\\
base & $v_x,v_y,\omega_z$; duration & Timed base velocity; actual motion is checked through subsequent observations.\\
stop & None & Stop base motion and cancel the joint segment, retaining applied joint targets.\\
finish & Outcome & End the attempt and prevent further motion.\\\bottomrule
\end{tabular}
\caption{Shared low-level operations. Joint angles use radians, base translation speed m/s, yaw speed rad/s, and durations seconds.}\label{tab:api}
\end{table}

Both interfaces cap base translation at 0.10\,m/s and yaw at 15 degrees/s; joint segments last at most 10\,s. Simulation base commands last at most 1\,s, whereas real base commands allow up to 5\,s of continuous output and can update the velocity without an intervening stop. Simulation observation triplets share one captured state; the three real USB cameras have individual acquisition timestamps. The real interface additionally provides \texttt{events}, \texttt{ack\_events}, and optional \texttt{wait\_feedback} operations for proactive operator reports. A success, wrong-button, or stop event immediately ends motion availability.

\subsection{Experience contents and pose provenance}
The simulation E3 package contains the Astra-generated summary, images at 22 sampled times, a timestamped image index and visualization, and a synchronized JSONL history linking requests to state or action feedback. The source episode is \texttt{be001-source-3}. E1 and E2 expose the corresponding subsets defined in Table~\ref{tab:design}; all use this same source. The real E3 package similarly combines the fastest A trial's summary, 25 image triplets, and timestamped action/state records. Reading prior images does not replace requesting fresh observations during a trial.

Figure~\ref{fig:arm-pose} uses recorded simulation configurations from \texttt{be001-source-3} at 0 and 481\,s and \texttt{be001-r1-i0e3} at 141\,s. The last two snapshots have measured shoulder, elbow, and wrist angles of approximately $(0.981,1.211,-0.837)$ rad, corresponding to the commanded $(1.0,1.2,-0.84)$ rad targets. Figure~\ref{fig:simulation-sequence} follows the latter trial at 0, 115, 150, 204, 281, and 296\,s, showing the approach and button activation before the trial ends at 310\,s. The release includes the selected configurations and button colors, original archive hashes, record indices, camera settings, and rendered images. Rendering restores recorded positions without advancing physics.

\subsection{Reproducing the figures and report}
The repository provides the simulation model, control and evaluation code, frozen condition inputs, trial-level results, selected videos, and reproduction instructions. Offline scripts rebuild the figures and tables from the released results, reproduce the prompt appendix from its frozen source files, and audit the behavior and timing diagnostics. Recorded-state renderers regenerate the simulation illustrations without running new agent trials. The self-contained manuscript source package includes the bibliography and all figure PDFs. New agent trials require compatible model access.
\end{document}